\documentclass[runningheads]{llncs}
\usepackage[portuguese]{babel}
\usepackage{graphicx}
\usepackage[utf8]{inputenc}
\usepackage{hyperref}
\usepackage{caption}
\usepackage{subcaption}
\usepackage{newunicodechar}
\newunicodechar{ﬁ}{fi}
\newunicodechar{ﬀ}{ff}

\begin{document}

\title{Deteção de estruturas permanentes a partir de dados de séries temporais Sentinel 1 e 2
\thanks{Trabalho financiado por fundos nacionais através da FCT – Fundação para a Ciência e a Tecnologia, I.P., no âmbito do projeto UID/CEC/04516/2019 (NOVA LINCS).}}

\author{André Neves\inst{1} \and
Carlos V.~Damásio\inst{1} \and
João M.~Pires\inst{1} \and 
Fernando Birra\inst{1}
}

\institute{NOVA-LINCS, DI, Faculdade de Ciências e Tecnologia, Universidade Nova de Lisboa \email{nova-lincs.secretariado@fct.unl.pt} \\ \url{http://nova-lincs.di.fct.unl.pt/}}
\authorrunning{A. Neves et al.}

\titlerunning{Deteção de estruturas permanentes a partir de dados de séries temporais}
 
\maketitle
\begin{abstract} 
Mapping structures such as settlements, roads, individual houses and any other types of artificial structures is of great importance for the analysis of urban growth, masking, image alignment and, especially in the studied use case, the definition of Fuel Management Networks (FGC), which protect buildings from forest fires.
Current cartography has a low generation frequency and their resolution may not be suitable for extracting small structures such as small settlements or roads, which may lack forest fire protection.
In this paper, we use time series data, extracted from Sentinel-1 and 2 constellations, over Santarém, Mação, to explore the detection of permanent structures at a resolution of 10 by 10 meters. For this purpose, a \textit{XGBoost} classification model is trained with 133 attributes extracted from the time series from all the bands, including normalized radiometric indices.
The results show that the use of time series data increases the accuracy of the extraction of permanent structures when compared using only static data, using multitemporal data also increases the number of detected roads.
In general, the final result has a permanent structure mapping with a higher resolution than state of the art settlement maps, small structures and roads are also more accurately represented. Regarding the use case, by using our final map for the creation of FGC it is possible to simplify and accelerate the process of delimitation of the official FGC.
\end{abstract}


\begin{abstract} 
O mapeamento de estruturas tais como aglomerados populacionais, estradas, casas individuais e qualquer outro tipo de estrutura artificial é de grande importância para a análise do crescimento urbano, criação de máscaras, alinhamento de imagens e, especialmente no caso de uso estudado, definição das faixas de gestão de combustível (FGC).
As cartas existentes possuem uma frequência de geração baixa e a sua resolução poderá não adequada para a extração de pequenas estruturas, como pequenos aglomerados ou estradas, que carecem de proteção contra incêndios.
Neste artigo, é explorada a utilização de séries temporais, provenientes dos satélites Sentinel-1 e 2 sobre o grânulo de Santarém, Mação, para uma deteção de estruturas permanentes a uma resolução de 10 por 10 metros. Para isso, é treinado um modelo \textit{XGBoost} com 133 atributos extraídos de todas as bandas das séries temporais, incluindo índices radiométricos normalizados.
Os resultados obtidos mostram que a utilização de séries temporais aumenta a precisão da extração de estruturas permanentes quando comparados com dados estáticos. A utilização de múltiplos dados de referência resulta num aumento do número de estradas detetadas.
Em geral, o resultado final possui um mapeamento de estruturas permanentes com uma resolução superior à das cartas de referência, em que as estruturas pequenas e estradas são representadas com maior exatidão. Com a sua utilização para a criação de FGC é possível simplificar e acelerar o processo da delimitação das FGC oficiais.
\end{abstract}


\section{Introdução}
Com o lançamento recente de vários satélites, como os satélites da constelação \textit{Sentinel-1} e \textit{Sentinel-2}, a disponibilidade de vários tipos de dados sobre a Terra têm vindo a aumentar, criando assim novas oportunidades que estão a ser exploradas para uma melhor monitorização terrestre.
Uma das áreas exploradas é a geração de cartas de ocupação do solo; este tipo de cartografia possui informação importante e é utilizada como base na construção de mapas de risco, criação de máscaras de zonas com estruturas permanentes, planeamento urbano e muitos outros. Entende-se por estrutura permanente qualquer estrutura constituída por materiais artificiais feitos pelo homem em que a sua posição, ou existência, não é alterada num dado período de tempo desde a sua criação. Esta definição engloba as classes de nível 1, 2 e 3, denominadas como territórios artificializados da carta de ocupação de solo de 2015 (COS)~\cite{Direcao-GeraldoTerritorio2018}.

Geralmente, este tipo de cartografia possui uma resolução espacial e temporal não adequada a algumas aplicações. Por exemplo a última versão da carta COS é de 2015 e apenas são cartografadas estruturas com área superior a 1 hectare. Abordagens automáticas de deteção surgem como resposta a estes problemas, utilizando dados com grande frequência de geração e resolução.

Um bom mapeamento de estruturas permanentes é também crítico para o planeamento do combate a incêndios. Em particular a monitorização e classificação de zonas urbanas ajudam no suporte a um bom desenvolvimento urbano, gestão de risco e de desastres \cite{Pesaresi2015} e delimitação de zonas de vigilância para prevenção de vítimas e danos em caso incêndio. 
De todas as aplicações das cartas de estruturas permanentes é destacada a definição das faixas de gestão de combustível. 
As faixas são criadas por várias entidades oficiais e, de acordo com os critérios utilizados, podem conter diferenças significativas nos seus limites. A utilização de cartografia de estruturas permanentes, quanto utilizado no processo, permite uma maior coerência na definição destas faixas.

Neste artigo, é explorado o potencial da combinação de dados de séries temporais do Sentinel-1 e 2, de 2016, a uma resolução de 10 por 10m para a deteção de estruturas permanentes, utilizando algoritmos de aprendizagem automática, destacando-se o \textit{Gradient Tree Boosting} pela sua performance durante o estudo.
O artigo explora também a utilização dos resultados da extração de estruturas para a definição das faixas de gestão de combustível. Este caso de uso é destacado pois pode acelerar e simplificar a definição das faixas pelos municípios.

O artigo está estruturado da seguinte maneira: 
\begin{itemize}
    \item \textbf{Secção 2}: estado da arte, onde são apresentados produtos existentes e estudos sobre a mesma área.
    \item  \textbf{Secção 3}: abordagem, nesta secção é apresentada a região de estudo bem como as fontes de dados utilizadas e o algoritmo de aprendizagem automática considerado.
    \item  \textbf{Secção 4}: apresenta os resultados obtidos e compara os resultados obtidos da classificação de estruturas permanentes com os produtos e estudos apresentados no estado de arte.
    \item  \textbf{Secção 5}: o potencial desta metodologia é discutido nesta secção, onde são apresentadas as conclusões e os trabalhos futuros a desenvolver.
\end{itemize}{}

\section{Estado da arte}
A missão Sentinel-1 é a primeira constelação de satélites do programa Copernicus, com o primeiro satélite lançado em abril de 2014. Estes satélites captam informação utilizando radar de abertura sintética (SAR), existindo vários estudos que investigam o seu potencial para mapeamento de ocupação do solo \cite{Abdikan2016,Balzter2015}.
Baltzer et al, 2015 \cite{Balzter2015} investigaram o uso de imagens do Sentinel-1 para mapeamento de cobertura terrestre. Utilizaram métodos de \textit{ensemble}, nomeadamente Random Forest. Os resultados obtidos mostram o potencial da utilização deste satélite para classificação de ocupação de solo, é concluído que a informação destes satélites pode complementar abordagens existentes no que toca a classificação de solo. O estudo refere também que a utilização de dados multi-temporais possuem o potencial de aumentar a performance das classificações.

A constelação Sentinel-2 \cite{Bertini2012} é um projeto conduzido pela agência espacial europeia, com o primeiro satélite lançado em junho de 2015.
A boa resolução destes satélites tornam-nos apelativos visto que este artigo é focado na deteção de estruturas permanentes, o que inclui pequenos aglomerados e estruturas individuais. 
O estudo de Pesaresi et al. \cite{Pesaresi2016} mostra que a utilização das bandas de grande resolução
de 10 por 10m do Sentinel-2 aumentam a qualidade da classificação de mapas de ocupação de solo devido ao detalhe espacial acrescido. A classificação foi desenvolvida utilizando um método de aprendizagem designado \textit{Symbolic Machine Learning}. Os autores concluíram que a classificação possui detalhe acrescido e contraria alguns dados de treino, aglomerados populacionais dispersos foram classificados como estruturas permanentes mesmo sem existirem nos dados de referência. Existe ainda uma omissão de algumas estruturas, sendo este fenómeno observado em zonas rurais e em aglomerados populacionais de baixa dimensão.

De acordo com um estudo por Clerici et al.~\cite{Clerici2017}, a sinergia entre dados das missões Sentinel-1 e 2 permite uma melhor classificação de ocupação de solo. Este estudo mostra que a integração de dados espectrais e texturais produzem resultados com acurácias bastante satisfatórias. Concluem também que, em semelhança ao estudo de Baltzer et al.~\cite{Balzter2015}, a utilização de dados de séries temporais tem potencial para promover uma melhor discriminação entre classes.

É de grande importância referir também dois projetos de classificação de solo, a carta COS \cite{Direcao-GeraldoTerritorio2018} de 2015 e a \textit{Global Human Settlement Layer} (GHSL)~\cite{ghsl}.
A carta COS é gerada pelo Instituto Geográfico Português, tendo uma unidade mínima cartográfica de 1 hectare, o que significa que estruturas com menor área não são cartografadas. Esta é a carta utilizada neste estudo como referência para o treino do modelo.
O projeto GHSL tem como âmbito fornecer métodos e sistemas para o mapeamento robusto e automático de áreas urbanas \cite{Pesaresi2016}. A carta GHSL Landsat é gerada utilizando \textit{Symbolic machine learning} treinando em dados Landsat 8, com uma resolução final de 38x38m será utilizada como comparação dos resultados obtidos.

\section{Abordagem}
Este estudo teve como como objetivo a avaliação do potencial da resolução, a sinergia do Sentinel-1 e 2 e a utilização de dados de séries temporais para o mapeamento de estruturas permanentes. Todas as imagens utilizadas possuem projeção EPSG:32629. É também estudado a aplicação do resultado à definição de faixas de gestão de combustível.

\begin{figure}[t]
\centering
\includegraphics[width=\textwidth]{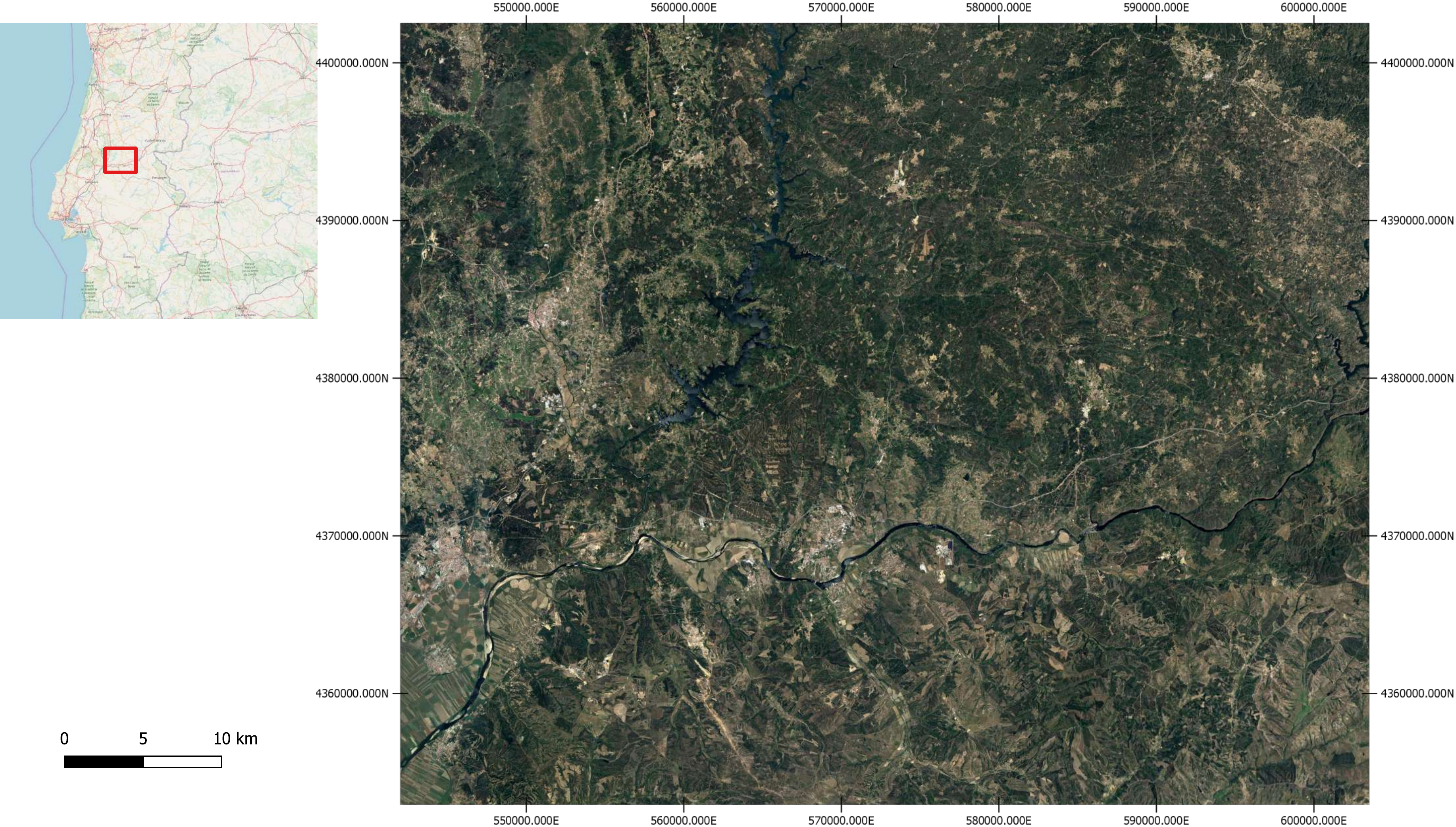}
\caption{Região de estudo, imagem do Sentinel-2 adquirida em Abril de 2016 sobre o distrito de Santarém.} \label{fig2:roi}
\end{figure}

Foi selecionada uma zona a Noroeste de Santarém, presente na Figura \ref{fig2:roi}, esta zona contém grandes aglomerados populacionais, assim como grandes superfícies com vegetação com vários aglomerados mais pequenos. Sendo maioritariamente composta por vegetação, por ser uma zona rural, esta zona apresenta um desafio aos algoritmos de aprendizagem automática.
O algoritmo selecionado neste artigo é \textit{Gradient Tree Boosting}, mais especificamente a ferramenta XGBoost \cite{Chen2016}, apesar de ser uma adição recente na classificação de solo \cite{Friedl2002}, esta metodologia é um alvo de estudo interessante. Este algoritmo tem uma grande vantagem sobre os algoritmos mais utilizados na classificação de solo, pois pode ser acelerado com GPUs, e logo o tempo de treino e de inferência é drasticamente mais baixo e os seus resultados rivalizam outros algoritmos mais usados \cite{Rogan2012}. 

A região, representada na Figura \ref{fig2:roi}, cobre uma área de 50x38 km o que, a uma resolução de 10 por 10m, equivale a aproximadamente 19 milhões de pixeis. Esta zona foi escolhida pela sua distribuição das faixas de gestão de combustível e por estar sobre apenas um grânulo de ambos os satélites Sentinel-1 e 2, o que simplifica e acelera o processo de processamento das imagens.

\subsection{Dados de referência}
\label{sub:refdata}
\begin{figure}
     \centering
     \begin{subfigure}[b]{\textwidth}
         \centering
         \includegraphics[width=\textwidth]{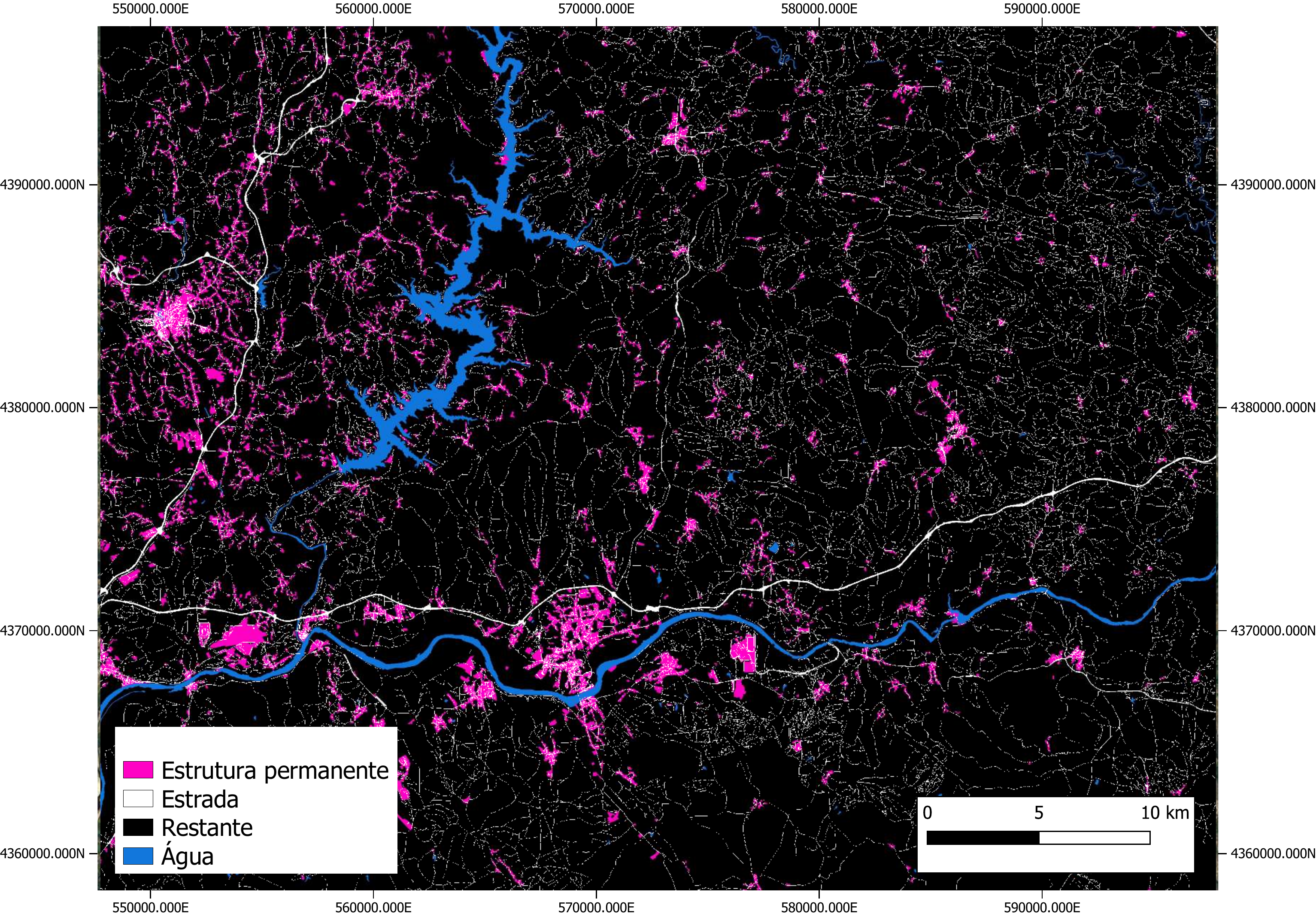}
         \caption{}
         \label{fig3:gt}
     \end{subfigure}
     \hfill
     \begin{subfigure}[b]{\textwidth} 
         \centering
         \includegraphics[width=\textwidth]{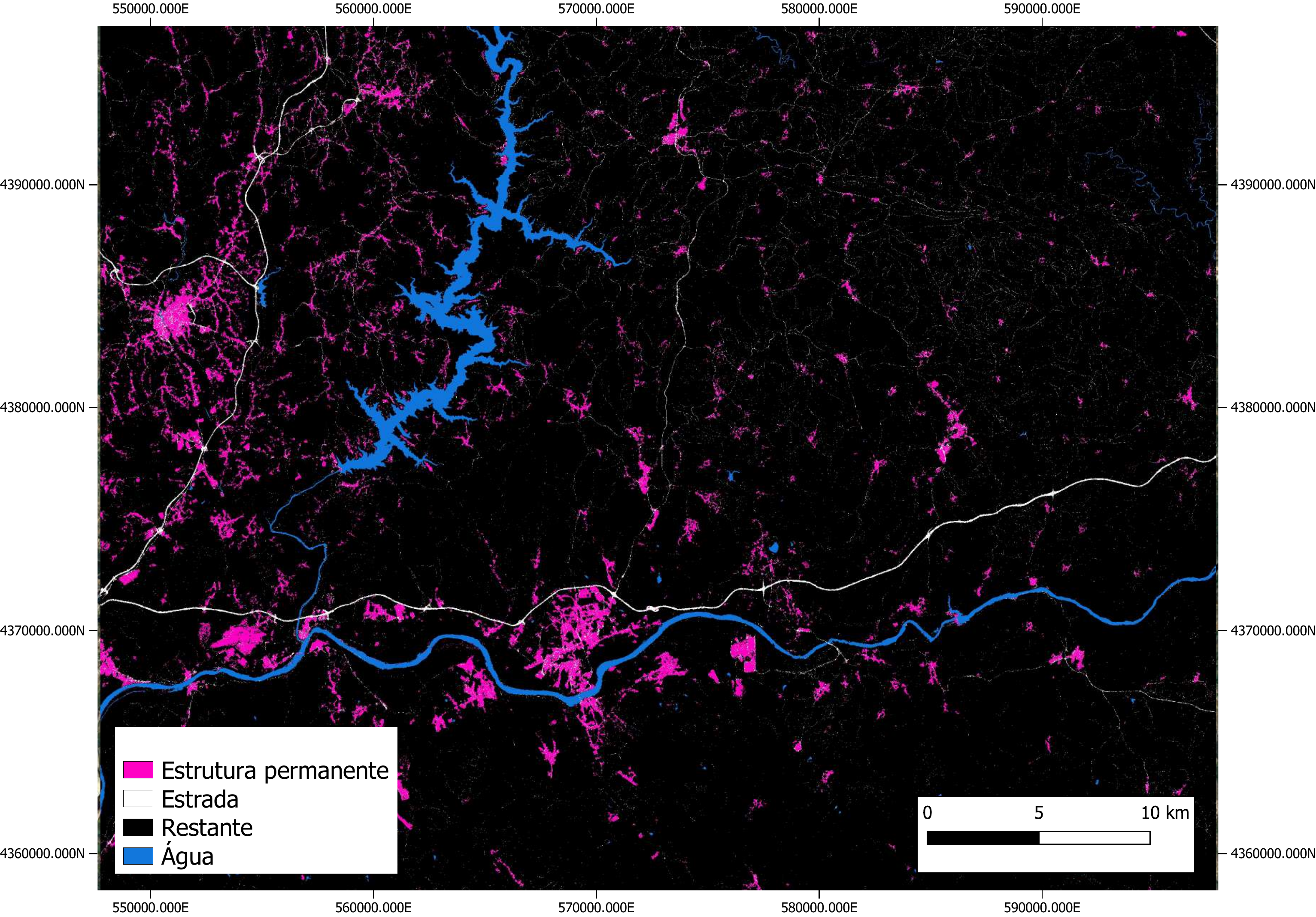}
         \caption{}
         \label{fig3:res}
     \end{subfigure}
    \caption{(a) Dados para aprendizagem, junção de estradas OpenStreetMap com a carta COS; (b) Resultado da classificação}
    \label{fig3:gtres}
\end{figure}

Para este estudo são consideradas 2 classes principais, ``Estruturas permanentes" e ``Estradas", com a adição da classe ``Água" devido a ser um caso especial das classes restantes e pode ser bem mapeada com esta abordagem.
Os dados de referência são extraídos da COS de 2015 \footnote{Direção-Geral do Território - \url{http://www.dgterritorio.pt}} e a informação sobre estradas é extraída do OpenStreetMap (OSM) \footnote{OSM - \url{https://www.openstreetmap.org}}. A classe água foi extraída das classes 4 e 5 de nível 1 do COS, a classe estrada foi extraída diretamente do mapa OSM e as estruturas permanentes extraídas da classe 1 de nível 1 do COS - territórios artificializados. As restantes classes são agregadas o que resulta na classe, ``Restante", representada a preto na Figura \ref{fig3:gt}.

A informação é então convertida em raster a uma resolução de 10 por 10m, os dados COS são complementados com os dados OSM (Figura \ref{fig3:gt}), substituindo zonas nos dados COS pela classe estradas do OSM.

Para a criação das estradas, visto que a informação proveniente do OSM é em formato de vetorial, foi necessário a criação de um \textit{buffer} dinâmico envolvendo as mesmas, sendo a largura deste \textit{buffer}, de entre 10 a 5 metros dependendo do tipo de estrada. 
Devido à possibilidade da existência estradas com o mesmo tipo mas com larguras diferentes e de estradas com tamanho inferior à resolução dos dados utilizados, irá existir algum erro durante a classificação, maioritariamente erros de omissão. 

\subsection{Recolha e tratamento de dados}
Os dados de Sentinel 2 foram extraídos com a ferramenta Copernicus API hub \footnote{Copernicus hub - \url{https://scihub.copernicus.eu}}, foram recolhidos dados sobre a zona de estudo com percentagem de nuvens até 30\%, as imagens mais contaminadas foram removidas manualmente, o que resulta em 18 produtos com 13 bandas cada, totalizando 252 imagens.
A partir destes dados são gerados os índices radiométricos normalizados, nomeadamente \textit{water index} (NWDI), \textit{vegetation index} (NDVI), \textit{built-up index} (NDBI) e o \textit{enhanced vegetacion index} (EVI).
O NWDI \cite{McFeeters1996} tem uma correlação negativa com áreas urbanas e superfícies impermeáveis \cite{Chen2006}, o NDVI \cite{Tucker1979} e a sua variante o EVI \cite{Huete2002} produzem índices espectrais que tem uma correlação negativa com áreas urbanas e o NDBI \cite{Zha2003} é usado para identificar áreas urbanizadas e tem uma correlação positiva com as mesmas bem como superfícies impermeáveis \cite{Chen2006}. 

A série temporal Sentinel-1 foi recolhida com o apoio do portal de dados da \textit{Alaska Satellite Facility} \footnote{Portal de dados do Alaska - \url{https://vertex.daac.asf.alaska.edu/}}, este mantém dados de 2016 sendo mais fácil a sua obtenção. No final foram obtidos 12 produtos com polarização dupla e uma órbita ascendente, processados da seguinte forma: \textit{Thermal Noise Removal} $\rightarrow$ \textit{Calibration} $\rightarrow$ \textit{Apply orbit file} $\rightarrow$ \textit{Terrain Flattening} $\rightarrow$ \textit{Terrain correction}\footnote{SNAP - ESA Sentinel Application Platform v2.0.2, \url{http://step.esa.int}}

Calculando a média, percentil 0, 25, 50, 75, 100 e a variância ao longo da série temporal sobre as bandas originais e índices a uma resolução de 10 por 10m, foram utilizados os seguintes atributos:
\begin{itemize}
    \item Atributos espetrais: todas as bandas do Sentinel 2.
    \item Índices radiométricos: NDVI, EVI, NDBI e NDWI, calculados a partir das bandas do Sentinel 2. 
    \item Atributos radar: polarização dupla do Sentinel-1, VV e VH.
\end{itemize}

\begin{figure}[t]
\includegraphics[width=\textwidth]{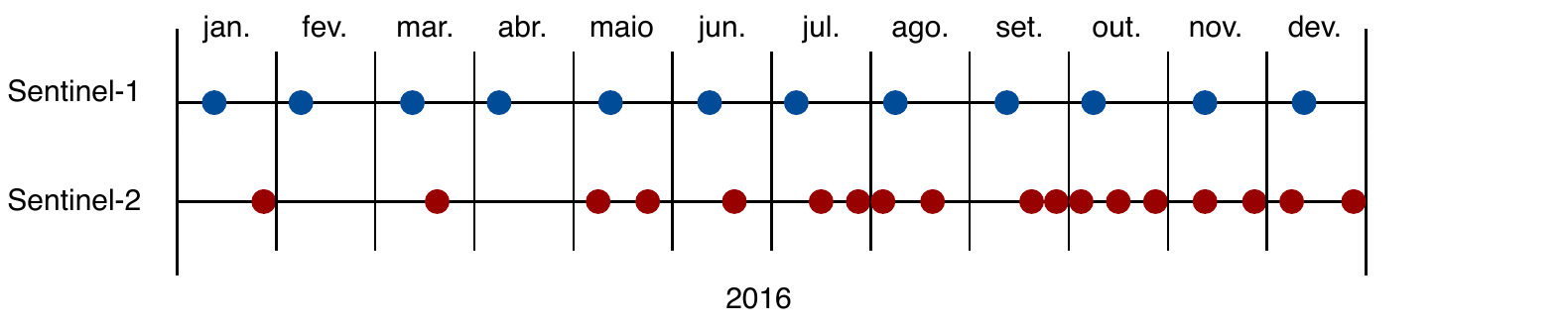}
\caption{Distribuição temporal das imagens recolhidas} \label{fig:data}
\end{figure}
A distribuição temporal das imagens recolhidas está representada na Figura \ref{fig:data}, existindo uma diferença temporal entre os dados de referência (2015) e os dados de treino usados (2016).
Durante a recolha, foi descoberto um pequeno desalinhamento nas imagens, ao utilizar uma estrada proveniente do OSM, todas as imagens foram alinhadas manualmente utilizando a mesma como eixo de referência.

\section{Resultados}
\subsubsection{Classificação} Os resultados foram obtidos após o treino do modelo XGBoost utilizando aproximadamente 4 milhões de amostras aleatórias, com a distribuição apresentada na coluna ``amostras" da Tabela \ref{tab1:class}. A classificação final, comparada com os dados de referência, está presentes na Figura \ref{fig3:gtres}. 
A Figura \ref{fig3:res} mostra detalhe acrescido quando comparado com os dados de referência (Figura \ref{fig3:gt}), apesar de algumas estruturas não estarem presentes nos dados de referência, pois possuem áreas inferiores a 1 hectare, o classificador consegue detetar as mesmas. Isto foi concluído também no estudo por Pesaresi et al~\cite{Pesaresi2016}, caso exista informação suficiente nos dados de Satélite que contrariem os de referência os classificadores conseguem gerar uma imagem com detalhe acrescido.

\begin{table}[bp]
\centering
\caption{Métricas da classificação nos dados de validação}\label{tab1:class} 
\begin{tabular}{|l|l|l|l|l|l|l|l|l|}
\hline
& \multicolumn{4}{l|}{Sem estradas OSM} & \multicolumn{4}{l|}{Com estradas OSM} \\
\cline{2-9}
classe &  precisão  &  recall & F1-score & amostras &  precisão  &  recall & F1-score  & amostras  \\
\hline
Estrutura  &   0.86  &    0.72   &   0.78 & 33,284 &   0.73  &    0.66   &   0.70 & 28,754\\
Estrada    &   0.92   &   0.79   &   0.85  & 2,830 &   0.66   &   0.13   &   0.22 & 32,557\\
Restante    &   0.99    &  0.99 &     0.99 & 723,932 &   0.95    &  0.99 &     0.97 & 698,808\\
Água    &   0.99    &  0.95  &    0.97   & 15,419 &   0.97    &  0.95  &    0.96 & 15,346 \\
\hline
Kappa & \multicolumn{4}{l|}{0.84} & \multicolumn{4}{l|}{0.63} \\
Tempo de treino & \multicolumn{4}{l|}{0:32:53} & \multicolumn{4}{l|}{0:31:28}  \\
\hline
\end{tabular}
\end{table}

\begin{figure}
    \centering
    \includegraphics[width=\textwidth]{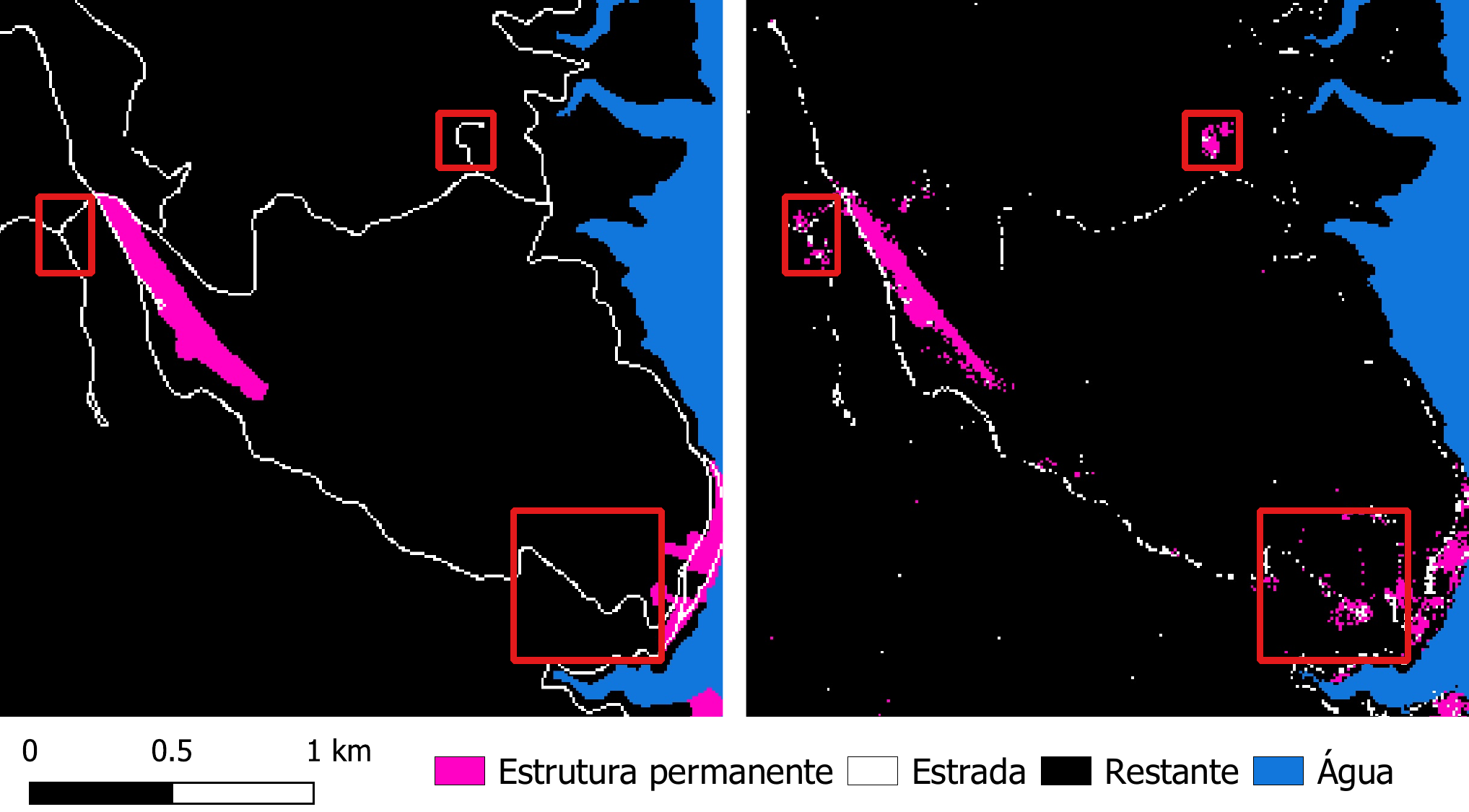}
    \caption{Dados para aprendizagem, junção das estradas OpenStreetMap com a carta COS, à esquerda; (b) Resultado da classificação, à direita}
    \label{fig4:gereseart}
\end{figure}
Na Figura \ref{fig4:gereseart}, assinalado a vermelho, podem ser observados pequenos aglomerados e estruturas não presentes nos dados de referência. Apesar deste comportamento ser benéfico, irá baixar as métricas de avaliação. 
O comportamento do classificador XGBoost é avaliado utilizando uma amostra aleatória de 775465 pontos. Comparado apenas com a carta COS, apresenta as métricas representadas na Tabela \ref{tab1:class}, chegando a valores de F1 e recall a cima dos 70\% e uma precisão de 86\%.
Quando comparado com os dados de referência, COS com OSM, as métrica são inferiores. Este comportamento é devido à adição das estradas, em que o seu recall desce de 79\% para 13\%. A descida das métricas relativas a estruturas permanentes é também consequência desta adição, naturalmente existem estradas em aglomerados populacionais, o que resulta numa confusão maior entre estradas e estruturas permanentes. Este erro tem pouco impacto no resultado final visto que as estradas são um subtipo de estrutura permanente e os aglomerados são facilmente detetáveis.
Apesar das métricas mais baixas os resultados mostram rivalizar os anteriores, devido à classificação de estradas mais pequenas que são inexistentes caso não se use dados de referência do OSM.

O Kappa \cite{Cohen1986} é uma estatística bastante utilizada na área de deteção remota~\cite{Inglada2017,Abdikan2016}. É introduzida neste estudo para combater a tendência dos algoritmos a favorecerem modelos com acurácia mais elevada, mas que apenas classificam bem uma classe. Os resultados obtidos possuem, respetivamente um Kappa de 0.84, comparando a carta COS, e 0.63, incluindo estradas OSM. Foi observado que existe uma diferença significativa, após a adição das estradas OSM, devido a um elevado número de erros de omissão das mesmas ($\approx70\%$). Algumas estruturas permanentes isoladas também não são incluídas no resultado, por serem mais pequenas do que a resolução de 10 metros dos dados utilizados, pois não possuem uma assinatura espetral suficiente para a sua deteção.

Para analisar a performance da utilização de dados dos satélites Sentinel-1 e 2 em simultâneo foram treinados dois modelos adicionais, um com a série temporal Sentinel-1 e outro com a série temporal Sentinel-2 apenas. Ao utilizar apenas dados Sentinel-1 o modelo obtém um Kappa de 0.37 e, com dados Sentinel 2, um Kappa de 0.59. Ambos os modelos são treinados com dados de referência COS mais estradas OSM.
Pode ser concluído que a série temporal Sentinel-2 tem mais peso na classificação de estruturas permanentes no caso estudado, esta possui 119 atributos, versos os 14 da série temporal Sentinel-1, o que faz com que exista mais informação presente nos dados Sentinel-2. Apesar do número reduzido de atributos da série temporal do Sentinel-1, quando utilizadas em conjunto, as duas séries temporais apresentam um resultado melhor.

\begin{figure}
    \centering
    \includegraphics[width=\textwidth]{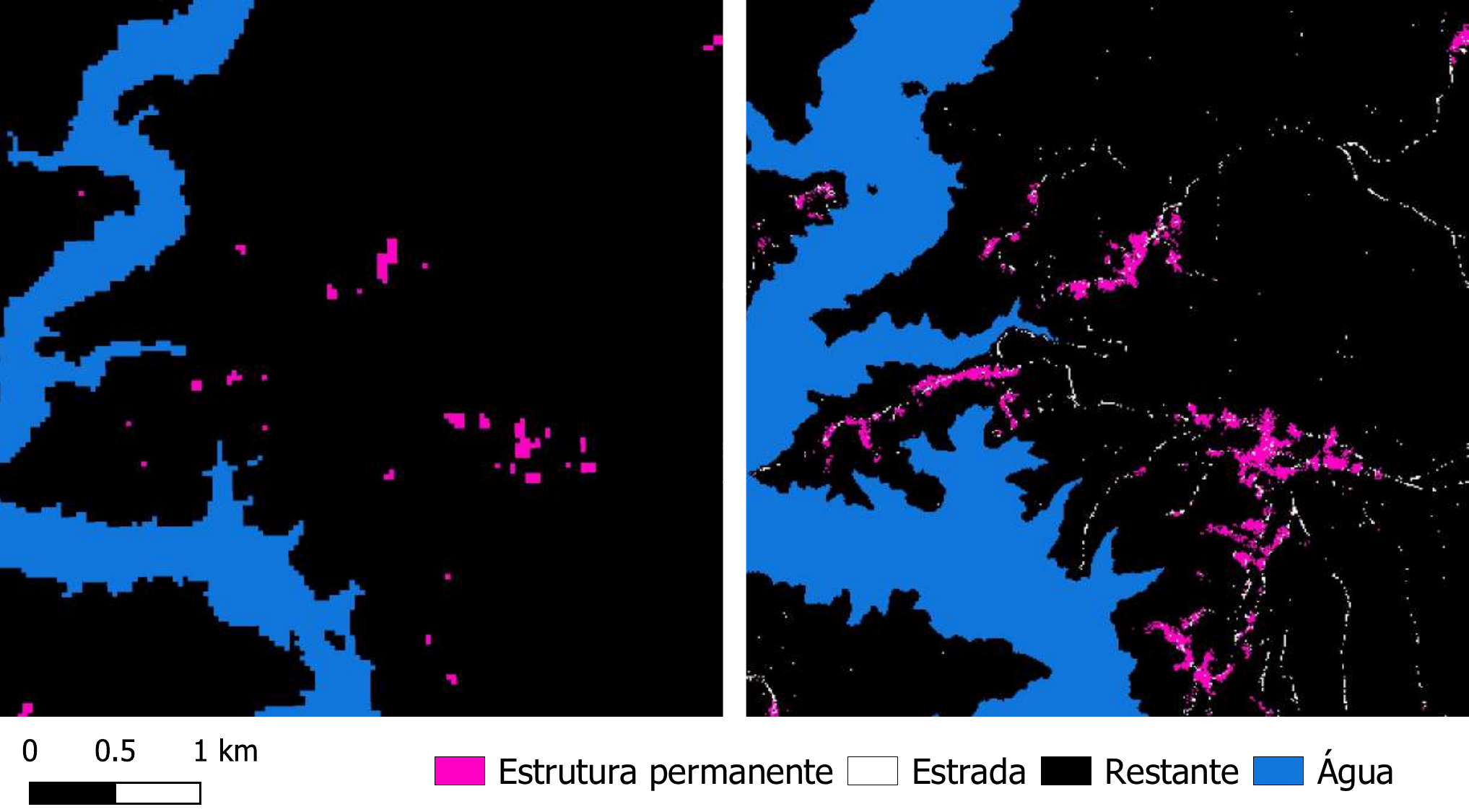}
    \caption{Global Human Settlement Layer de 2015, à esquerda; Resultado da classificação XGBoost, à direita}
    \label{fig5:res}
\end{figure}
Uma comparação da performance do classificador XGBoost com a carta (GHSL)~\cite{ghsl} de 2015 está presente na Figura \ref{fig5:res}, o resultado obtido possui detalhe acrescido, não só em pequenos aglomerados e estruturas isoladas, mas também mapeia com mais precisão água, principalmente rios e corpos de água mais estreitos. Quando comparado com a carta COS sem informações de estradas, pois não as classifica separadamente, a GHSL obtém um Kappa de 0.46 e precisão de estruturas de 0.65, enquanto o resultado obtido possui ambas as métricas de 0.82.

\begin{figure}
\centering
\includegraphics[width=\textwidth]{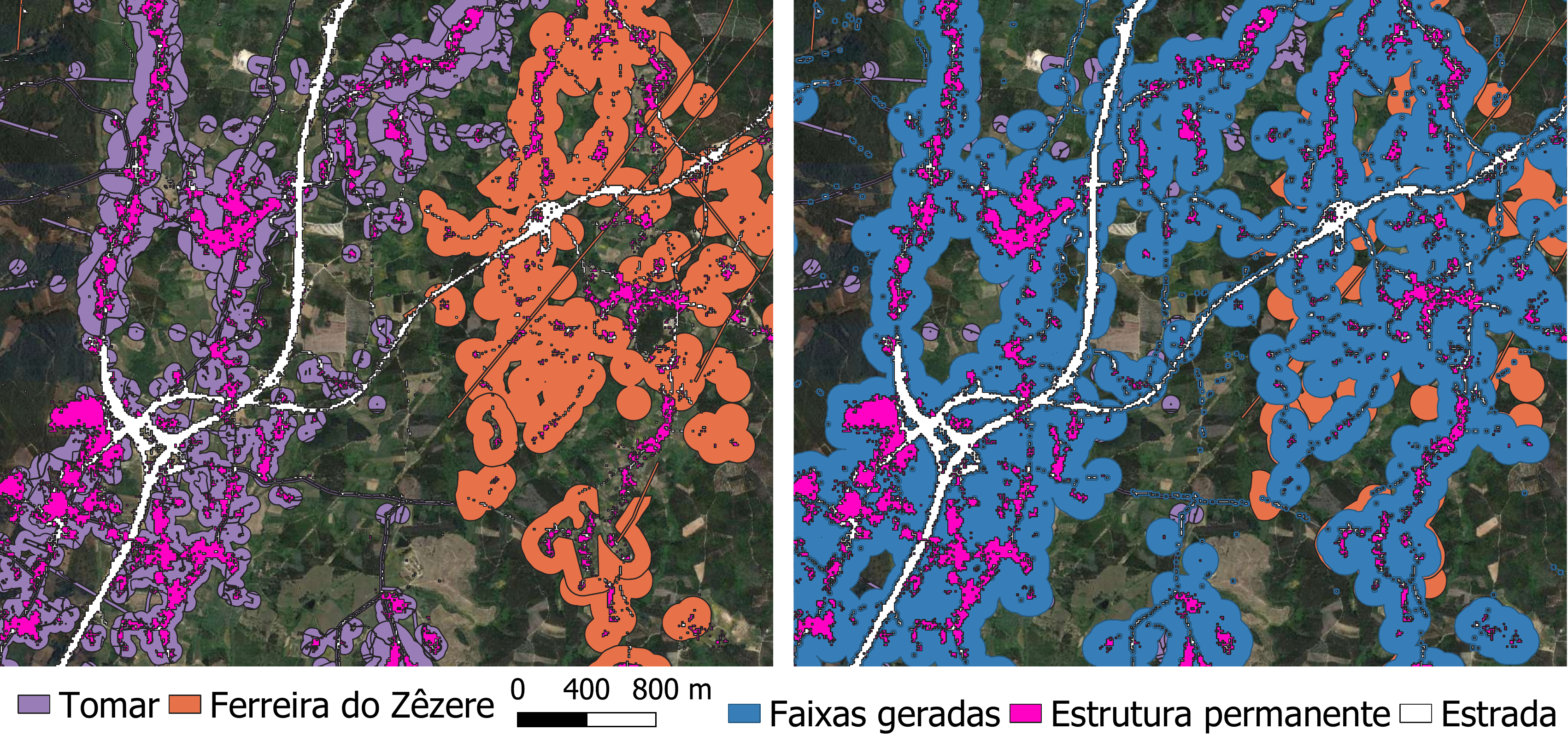}
\caption{FGC oficiais de Tomar e Ferreira do Zêzere, à esquerda; FGC definidas a partir da classificação, à direita} \label{fig6:fgcred}
\end{figure}
\subsubsection{Resultados FGC} A Figura \ref{fig6:fgcred} apresenta o resultado do algoritmo XGBoost comparado com as faixas, obtido após aplicar um \textit{buffer} com largura de 100 metros. Observa-se que os resultados mapeiam quase perfeitamente as faixas de Tomar, indicando uma robustez dos resultados obtidos. Quando sobreposto às faixas de Ferreira do Zêzere é possível identificar várias estruturas e aglomerados populacionais sem as respetivas faixas nos dados oficiais, sendo essencialmente observado em zonas de menor densidade ou estruturas singulares.

\section{Conclusão}
    A utilização de dados de séries temporais do Sentinel-1 e 2, mostram uma sinergia entre os dois satélites para a deteção de estruturas permanentes. O comportamento destas estruturas ao longo do tempo e a sua interação com o ambiente difere suficientemente das restantes classes para uma boa classificação. Esta classificação demonstra ser adequada a várias aplicações, como gestão de risco e de desastres \cite{Pesaresi2015}, criação de máscaras de estruturas, alinhamento de imagens e mapeamento de populações \cite{Freire2015} entre outras.
    
    A classificação final do estudo, apresenta métricas satisfatórias. Quando comparado com a carta GHSL, o resultado possui detalhe acrescido, são detetados aglomerados populacionais mais pequenos e estruturas mais estreitas, como as margens de rios e estradas. Não será possível comparar as métricas do resultado diretamente com os estudos apresentados no estado da arte, devido à utilização de dados de referência diferentes. A falta de disponibilidade dos dados ou algoritmos também impossibilita uma comparação visual, embora isto abra uma oportunidade para um estudo futuro do desempenho da metodologia apresentada com outros trabalhos que disponibilizem os seus recursos.
    
    Os resultados apresentados neste artigo mostram potencial para a validação, ou até na definição de faixas de gestão de combustível, assistida ou totalmente automática. A aplicação desta metodologia ao processo de definição das faixas promove a proteção das populações, simplificando o processo e aumentando a coerência dos seus resultados.
    
    Futuramente, será possível melhorar os resultados utilizando dados de referência mais refinados. Estes devem incluir edifícios singulares e não apenas aglomerados, o que tira partido da grande resolução dos dados utilizados. Com o aumento dos recursos disponíveis será possível também adicionar análise de textura às bandas do Sentinel-1, o que tem demonstrado bom potencial na classificação de solo \cite{Balzter2015,Clerici2017}. 
    Para melhorar a classificação está a ser estudada a possibilidade da aplicação de algoritmos de convolução, estes modelos atingem resultados elevados no que toca à classificação de imagens \cite{Fu2017}.
   
\bibliographystyle{bibstyle}
\bibliography{inforum2019paper}
\end{document}